*Я.А. Перванов (Италия)*

**МАТЕРИАЛЫ К РУССКО-БОЛГАРСКОМУ СОПОСТАВИТЕЛЬНОМУ СЛОВАРЮ «СЭД»**

> Итак, мы убеждаемся в необходимости отражения в двуязычном словаре не отдельных слов и их иноязычных эквивалентов, а целых фрагментов лексических систем двух языков, представленных в сопоставительном плане. Сложность создания словарной статьи такого типа очевидна, поэтому до сих пор образцов словарей, построенных по такому плану, не существует. Кроме того, в словаре с такой организацией возникнет много новых проблем, начиная от порядка следования лемм и их поиска, заканчивая объемом [Паликова 2007: 67].

*Введение*

Цель настоящей статьи – предложить один из путей приближения к реальному прототипу сопоставительного электронного русско-болгарского словаря нового образца, в котором накопленные знания о сценариях говорящего и наивной «картине мира» (схемы, фреймы, гештальты) были бы надлежащим образом синхронизированы с лексикографической интерпретацией значений и научно обоснованным сопоставлением двух или более языков.

Очевидно, создание подобного словаря сопряжено с необходимостью ответить на ряд вопросов, среди которых центральное место занимают вопросы о структуре, методе описания и предназначении словаря. Несомненно, такой словарь невозможно создать в виде большого проекта, без предварительной проработки экспериментального словаря небольшого размера, проверки его функциональных возможностей, технических решений, опроса читателей и маркетинга.

К настоящему времени одна лишь идея создания сопоставительного (не просто переводного) словаря могла бы обескуражить многих. Если в переводном словаре по традиции задано направление: от неродного к родному или наоборот, то в сопоставительном словаре понятие «родного» терминологически бесполезно, а направление как таковое – теряет свою «переводческую» подоплеку. Если, например, в двуязычном словаре болг. *заблуждавам* имеет эквиваленты *обманывать, вводить в заблуждение,* и на этом сопоставление заканчивается, то в сопоставительном словаре – это лишь начало, поскольку необходимо определить, во-первых, симметричный или асимметричный характер соотношения эквивалентов на фоне других эквивалентов типа *обманывать – лъжа, лъжа – лгать, вводить в заблуждение – вкарвам в заблуда* и т.д., во-вторых, представить все эти эквиваленты в упорядоченной классификации, в третьих, найти новый способ лемматизации материала (пар) в словарной статье. Это лишь некоторые различия. В сопоставительном словаре, в отличие от переводного, необходимо отражать толкования значений многозначных слов, а не полагаться на интуицию читателя или лингвиста. Далее, необходимо включить обширные цитаты из национальных корпусов, было бы неплохо учесть синонимы, фразеологизмы и перифразы…

Ситуация может оказаться намного сложнее, если задаться целью не просто сопоставлять по алфавитному порядку произвольные группы слов, а включить идею антропоцентризма в сопоставительное описание лексики двух языков. В таком случае словарь неизбежно приобретет элементы идеографического или семантического словаря, но придется также подумать об объеме словаря, о подготовленности читателя, а главное – коренным образом изменить макроструктуру словаря и сделать его интерактивным.

Предлагаемый Словарь имен экспансивных действий (далее – СЭД), действующий прототип которого опубликован на сайте автора, базируется на результатах сопоставительного исследования наименований экспансивных действий и признаков в русском и болгарском языках [Перванов 1995, 2009]. Это очень интересная часть лексики и богатая почва не только для лингвиста. Своего рода «аттрактивная агрессорша». Экспансия – это целенаправленные действия X-а (совокупности X-ов) либо состояния X-а, которые или по характеру, или по замыслу, или по результатам способны вызвать изменения в сфере Y-а (совокупностиY-ов), изменить привычный для него статус, норму, создать ситуацию дисбаланса интересов, сил, возможностей, но никогда – в пользу толькоY-а. В общем случае экспансия проявляется как влияние, вмешательство в некоторое положение дел на стороне Y-а, поэтому она крепко спаяна с аксиологией говорящего.

Под типом экспансивного действия мы подразумеваем наиболее общие логические характеристики действия, устанавливаемые на базе сопоставления и сближения значений глаголов и определяемые с точки зрения того, каков основной или вторичный результат действия, какие изменения в сфере Y-а вызывает это действие, каковы цели действия. Типология экспансивных действий была разработана автором применительно к лексическому массиву русского языка. Выделены и обстоятельно классифицированы следующие типы экспансивных действий: ликвидирующие действия, деформирующие действия, аннексирующие действия, дезориентирующие действия, принижающие действия, блокирующие действия, действия вмешательства, провоцирующие действия, регулирующие действия, экспансивное поведение.

Классификация лексического материала в идеале должна учитывать все имена, могущие обозначать экспансию, экспансивные качества субъекта X и онтологический результат его влияния на Y-а. Это значило бы, что при объеме однотомного словаря в 50000 слов не менее 2000 из них были бы каким-то образом втянуты в исследование, включая существительные, прилагательные, глаголы и наречия. Количество исходного материала существенно возрастет при включении фразеологизмов, а также с учетом перифраз, коннотированной, инвективной лексики и синтагматически обусловленных значений слов (ср.: *отдать книгу* и *отдать распоряжение*, *восстановить в должности* и *восстановить против кого-либо*, *привлечь внимание* и *привлечь к ответственности*, *ахнуть* и *ахнуть кого-либо* 'ударить', *просьба* и *категорическая просьба, инициатор* и *застрельщик, солить* и *насолить* 'причинить неприятность' и т.д.).

Оказывается, что мы имеем дело не с узкой тематической группой, а с целым корпусом слов, который заслуживает своего словаря. Нами было высказано также предположение, что данная макрогруппа словаря не является чисто языковым феноменом, а опирается на тезаурусную схему экспансивного действия, или «подъязык» говорящего. Схема, в отличие от языка, не членится на лексемы и значения, в ней выделяются лишь значимые фрагменты.

До разработки СЭД мы попытались ответить на ряд вопросов:

Каковы составляющие экспансивного действия? Чем оно отличается от других действий? Достаточны ли логические критерии для его характеризации, существует ли чисто концептуальный каркас семантики экспансивного действия? Содержатся ли в средствах языка жесткие указания на то, что действие экспансивно? Какова «дисперсия» этой тезаурусной схемы на уровне лексического обозначения, или каковы вербальные модификации схемы? Как распределяются семантические роли в экспликации схемы, отдельных ее фрагментов между языковыми концептами? «Экспансивна» ли схема этого действия и как это отражается на уровне двух родственных языков в плане привлечения сверхнормативных средств обозначения [Перванов 2010]?

С учетом упомянутых различий между переводным и сопоставительным словарем и принимая во внимание переплетение интралингвальных и экстралингвальных факторов, формирующих значение слова, мы вводим в словарь некоторые новые понятия и термины. Отметим, что при всей небезупречности они в значительной степени предопределяют совмещенное видение лексики двух языков, избавляют от «челночных» процедур сопоставления и позволяют организовать страницы словаря по строгим правилам.

Итак, попытаемся построить изложение концепции сопоставительного электронного словаря как краткий ответ на известные восемь вопросов Ч. Филлмора [Филлмор 1983: 23 – 24].

## 1. Каков способ представления информации в словаре?

В сопоставительном электронном словаре русского и болгарского языков нет традиционной словарной статьи. Словарная статья является страницей. Внешне страница похожа на диаграмму, по сути является «множеством зеркал», по перцептивному признаку – моментальным снимком семантического поля, обозначающего данный фрагмент тезауруса. Семантическая информация скрыта до момента нажатия мышью на слово или другой вспомогательный элемент. В гиперпространстве словаря страница имеет свою индивидуальность – заголовок (заголовочная пара). Страница состоит из трех значимых компонентов: с л о в а (лексические эквиваленты не меньшие, чем слово), компонованные в «двоичные знаки», и д е о г р а м м ы, отображающие классификацию эквивалентов в сопоставительной модели словаря, их соотношение, с п р а в о ч н а я   ч а с т ь  в роли контекстуальной помощи читателю.

Формат страницы – конвертируемый электронный формат HTML, XML.

Лексическая информация страницы включает согласованное двуязычное толкование значений членов пар, корреляцию пар, синонимию, выборки из национальных корпусов, фразеологические соответствия, классификацию значений по принадлежности к определенному фрагменту тезауруса, индекс результата экспансивного действия и др.

Представление этой информации следует принципам простоты, эргономики и эстетики страницы: 1. Компактность: никаких лишних прокруток страницы вверх-вниз. 2. Локализация: всплывающие окна (pop-ups) контекстуально привязаны к уровню пары. 3. Самодостаточность страницы: минимизация пересылок к другим страницам и возможных при этом ошибок в навигации. 4. Цветовая гамма: не более 10 – 12 оттенков. 5. Концептограммы (идеограммы): образная классификация двоичных знаков. 6. Построчный характер чтения: каждая строка расширяет семантическое поле словаря и стратегию пользователя. 7. Иррадиация лексической информации на строке – от центра к периферии. 8. Подключение к модулям в пределах страницы. 9. Прямые ссылки на источники в сети Интернет. 10. Комбинирование «книжного» и web-интерфейса и др. Подробнее об этом ниже.

Преобладающая масса электронных двуязычных словарей имеет композицию, повторяющую (или копирующую, иногда пиратским образом) структуру словарной статьи печатного издания. В настоящем словаре, как мы отметили, нет традиционной словарной статьи. Это поначалу может показаться странным, непонятным и сложным. На самом деле идея проста. Страница словаря включает эквивалентные пары слов на каждой строке. Заголовочная пара вынесена в первую строку. Отношения слов внутри пары классифицированы и представлены идеограммой пары.

Многозначные слова образуют больше одной пары, т.е. имеют соположенные пары, которые тоже классифицированы. Не все пары слов являются переводными эквивалентами, есть ложные пары – омонимы. Они тоже классифицированы. Пара слов определяется по значениям членов пары, толкование значений всплывает при нажатии мышью на слово. Страница читается построчно, каждая строка добавляет новые пары и расширяет смысловое поле заголовочной пары. Электронная композиция страницы позволяет в любой момент получить справку под чертой либо подробное объяснение в основном модуле словаря.

Пара сопоставимых слов в двух языках называется «двоичным знаком». Этот условный термин нуждается в пояснении. Основное различие между понятиями «лексическая пара», «лексический эквивалент», «коррелят» и т.д. и предлагаемым термином состоит в его классифицирующей функции: двоичный знак в СЭД – это пара (слов) с  п р е д в а р и т е л ь н о  о п р е д е л е н н ы м  типом соотношения, которая противопоставлена другим парам с иным типом эквивалентной связи. Модель сопоставления является одновременно моделью переводных эквивалентов и классификацией соотношений слов в двух языках. Так, болг. *лъжа* и рус. *лгать, врать, обманывать, вводить в заблуждение, брехать* входят в отношение эквивалентности, которое является уникальным для каждой пары и должно быть классифицировано. Двоичность в предлагаемом контексте несет на себе также отпечаток феноменов генетической общности и идиоматичности русского и болгарского языков.

### 2. Связь лексических единиц друг с другом. Что изменится при исключении из списка одной единицы или введении в него новой единицы?

Сопоставительный словарь чувствителен к принципу нормативности в отборе слов. Проблема не только в степени полноты словаря. Исключение определенного слова из списка сопоставляемых лишает нас возможности прогноза ошибки, например, рус. *лажа* и болг. сущ. *лъжа*, гл. *лъжа*. Включение нового «необкатанного» слова типа рус. *экзекутировать* (в омонимичных значениях 'казнить' и 'исполнить команду execute в электронном приложении') кардинально меняет структуру словарной страницы КАЗНИТЬ – ЕКЗЕКУТИРАМ.

В предлагаемом словаре ядерную группу составляют имена экспансивных действий. Следует ли рассматривать рус. *окружать* (*деревья окружали дом*) и *окружить* (*войска окружили деревню*) как два разных слова или определять значение СВ как привязанное к фрагменту блокирующих действий в тезаурусной схеме?

### 3. В какие отношения вступают лексические единицы между собой?

Классификация двоичных знаков в словаре определяет не только приоритет пар, но и семантические пропорции между языками. Эти пропорции отливаются в типы знаков, находящиеся в перекрестной связи. На страницах словаря представлены импликативные цепи знаков, которые в своей совокупности отображают живую динамику семантических полей в двух языках. Подробнее об этом ниже.

### 4. Какие виды стратификации следует признать в словарном составе языка?

Стилистически маркированные слова в двух языках, оценочная лексика и слова, обозначающие реалии, являются камнем преткновения для лексикографа. Сопоставительный словарь имен экспансивных действий не является исключением. Приведем пример: р. *врать*, *лгать* – б. *лъжа* (отныне для пар слов будем использовать цветовую гамму словаря – Я.П.). Безусловно, пара лгать лъжа гомогенна, нейтральна, и может оформить страницу в словаре. Однако по данным русского ассоциативного словаря [РАСС: электронный ресурс] в перекрестной выборке реакций «врать – лгать – обманывать» безусловен приоритет разг. *врать* по количеству реакций и широте оценок: *врать* (23), *лгать* (5), *обманывать* (4). Очевидно, в словаре должна быть страница «врать – лъжа», а пара может быть определена как синхронный гетерогенный знак.

### 5. Как можно убедиться в том, что определенный способ организации информации, наблюдаемый нами в том или ином описании словарного состава, отражает реальности самого языка, а не является следствием какой-то конкретной методологической установки?
### 6. Какие типы понятий или смысловых структур лексикализуются в языке, то есть обретают жизнь в виде самостоятельных лексических единиц? Какого рода описание было бы здесь наиболее предпочтительным?
### 7. Как распределяются семантические свойства между различными классами слов? Как меняется такое распределение от языка к языку?

Применительно к СЭД проверка на субъективность заключается в использовании данных других словарей. Методологической установкой является принцип «разрушения», обоснованный В. Скаличкой: всякое сравнение двух языков отчасти означает некоторое разрушение двух систем, ибо сравниваются какие-то части, подчас занимающие в этих системах совсем разные места [Скаличка 1989].

В сопоставительном словаре первым правилом представления двух языков является не переводимость с одного языка на другой, а паритетность обоих языков, их равенство перед метаязыком описания – концепцией совмещенного знака. Этот общий принцип гласит: отношение между семантикой фрагмента тезауруса носителя языка и (группирующимися в лексико-семантические поля) средствами языка А является отношением экспликации определенной тезаурусной схемы или ее отдельного фрагмента. В этом отношении выражается специфика языка А. Обозначим эту связь как вертикальную. Близкородственные языки А и В (в данном случае - русский и болгарский) сохраняют эту вертикальную связь в генетическом плане (зона сходств, обусловленных генетическим родством). Перестановка семасиологических акцентов, вхождение иноязычной лексики, угасание единых для А и В ассоциаций между формами и значениями приводят к ситуации существования схемы в двух «ипостасях»: отношение различия между языками является вместе с тем и отношением двоякого обозначения фрагментов схемы. Эта межъязыковая ситуация может быть обозначена термином «семантическая поляризация фрагментов схемы», а эквивалентные слова А и В могут быть развернуты в ряды «двоичных знаков». Отношения этого типа являются горизонтальными и могут быть классифицированы.

*8. Каковы различия между языками с точки зрения структуры их словарного состава?*

Различия между способами втягивания слов в экспансивную парадигму, отмечаемые для русского и болгарского языков, свидетельствуют о том, что семантические признаки этих слов имеют неодинаковую ценность с точки зрения внутрисистемных отношений передачи экспансии. Так, для русского языка более продуктивна модель аннексирующего действия, выражаемая глаголами с *обез-* (*обезлошадить, обезденежить, обезземелить, обезволить, обездолить* и др.). С другой стороны, для него не характерны некоторые способы совмещения каузации и результата в переходном глаголе типа болгарских *заблуждавам* 'вводить в заблуждение', *отчайвам* 'вводить в отчаяние', *наскърбявам* букв. 'делать скорбящим', *онеправдавам* букв. 'причинять кому-то неправду'. Русские *заблуждаться, отчаиваться, скорбеть* имеют только субъектную направленность, хотя и связанную с глубинным пациенсом.

При дезориентирующих действиях типа <обмерить, обвесить, обсчитать> прямой объект действия в русском связывается не с материалом (*ткань, колбаса, деньги*), а с тем, кто получит этот материал в меньшем, чем полагается, количестве. В болгарском отсутствует синестезия значений объекта действия и реципиента в одном предикате. Она может быть выражена либо двучленной структурой, включающей гностический результат 'лъжа, мамя' и обозначение действия <премервам, претеглям, броя по-малко>, либо идиомой типа *бия в кантара* букв. 'стукнуть по весам' [Русско-болгарский словарь – РБР].

Другие особенности: префиксальные спецификаторы экспансивного действия и степень продуктивности словообразовательных моделей в двух языках. Здесь можно прежде всего отметить функциональную значимость вариантов *пре-* и *пере-* для пар глаголов типа *предать – передать, преступить – переступить, преградить – перегородить* и т.д, нерелевантную для болгарского. С другой стороны, сравним б. *подигравам (се)* и р. *подыгрывать*, где словообразовательные модели расходятся с типизацией действий, в силу чего *подигравам* может образовывать только дизъюнктивные гетерогенные пары с русскими глаголами *подшучивать, издеваться (над кем-либо)* и др. (принижающие действия); б. *зарязвам* 'бросить, оставить' (сепарирующее действие) и р. *зарезать* – СВ к ликвидирующему *резать*. Что касается семантической роли приставки *о-/об-* и взаимосвязи русских производных *о-/об-* глаголов со схемой экспансивного действия, мы позволим себе цитировать содержательное исследование А.Д. Кошелева [Кошелев 2004].

Неодинаково соотносятся со схемой широкозначные слова типа рус. *брать – взять* и болг. *хващам, вземам*. Свойства глаголов входить в открытые ряды сочетаний предполагают тщательный анализ комбинаторики значений, вариативности, обусловленной динамикой таксономических и тематических классов актантов [Розина 2003].

*Место словаря в сложившейся лексикографической ситуации*

Мы связываем наши усилия по разработке сопоставительного словаря с некоторыми новыми проектами в лингвистике и лексикографии. Прежде всего, отметим выдвинутый А. Липовской проект комплексного электронного русско-болгарского словаря [Липовска 2006; Липовска 2009]. Предложенная А. Липовской концепция модульного словаря, выдвинутый принцип совмещения полей и способ «засвечивания» аналогов, на наш взгляд, имеют большую перспективу. А. Липовска – один из первых болгарских лексикографов, оценивших по достоинству перспективы электронного ословаривания нового типа на базе существующих данных переводных словарей и достижений сопоставительной лингвистики, ее идеи концептуально близки идеям СЭД по многим параметрам.

Цель проекта FrameNet под руководством Ч. Филлмора и К. Бейкера [Fillmore, Baker, Sato 2002] – документирование диапазона семантических и синтаксических комбинаторных возможностей (валентности) каждого слова в каждом из его значений, через машинную аннотацию примеров из языкового корпуса путем автоматической табуляции и показа результатов. Отличительная черта проекта – его фреймовая структура, включающая когнитивные элементы [FrameNet 2000: электронный ресурс].

Проект «Фрагменти от езиковата картина през призмата на вторичното назоваване» под руководством П. Легурской [БЕЛБ 2010: электронный ресурс] реализует синхронное сопоставление семантической структуры части предметной лексики на материале родственных и неродственных языков по инвариантной матричной модели с предсказуемой регулярностью вторичного означивания. По словам П. Легурской, вторичные значения «являются (перевод наш – Я.П.) точкой пересечения фоновых, экстралингвальных и лингвистических знаний носителей языка. В этой точке осуществляется переход от структурной семантики как метода исследования к когнитивному методу» [Легурска 2002: 51].

Принцип зеркального подобия реализуется в проекте Двуязычного экологического словаря-тезауруса [Табанакова, Ковязина 2007]. Концепция моделируемого двуязычного тезауруса предполагает разработку экологической терминосистемы в английском и русском языках изолированно друг от друга. «С точки зрения лингводидактического подхода словарь бифункционален, поскольку содержит вводную часть, дефиниции и пометы на двух языках, а также англо-русский и русско-английский алфавитные указатели экологических терминов. Следовательно, пользователем моделируемого словаря может быть как носитель русского, так и носитель английского языка» [там же, с. 31].

Первая и основная задача словаря СЭД – научно обоснованное сопоставление. Сопоставительный инструментарий словаря включает классификацию лексических параллелей между языками с учетом единого тезаурусного фонда – общей для обоих языков схемы картирования действительности под тем или иным углом зрения носителей языка. Эта часть словаря была бы интересна для специалистов по сопоставительной и когнитивной лингвистике.

Настоящий проект словаря вводит новый интегральный принцип – совмещение научного и прикладного в интуитивно понятной электронной графической оболочке. Именно это решение определяет практическую пользу и круг читателей словаря. В страницы словаря вкраплены идеограммы и частично интегрированы ссылки на словарные статьи из других словарей: семантических, идеографических, толковых, двуязычных и т.д. Можно сказать, что потенциал одной единственной электронной страницы словаря несравним с той информацией, которую могут дать в отдельности словарные статьи в толковых, синонимических и двуязычных словарях. Импликативные цепи слов включают согласованное двуязычное толкование для каждой пары, что очень важно для переводчика или преподавателя языка. Страницы настоящего словаря биэквивалентны и отображают лексические поля во взаимосвязи, в динамике и многообразии отношений многозначных слов в двух языках. Каждая страница является многослойной структурой и представляет собой моментный «снимок» двух совмещенных фрагментов тезауруса.

Словарь является словарем активного типа. Мы постарались придать страницам словаря электронную композицию, в которой каждый элемент взаимодействует с пользователем, и благодаря этому взаимодействию статика словарной статьи превращается в действующую модель Писателя. Обилие примеров, взятых из электронных корпусов, и комментарии к ним призваны помочь читателю, особенно в случаях, когда толкование слова не дает исчерпывающего ответа на ряд вопросов либо возникают сомнения относительно точности самого толкования. В словаре каждое слово имеет не только лексикографическую интерпретацию (читателю было бы интересно сопоставить толкования одного и того же значения в разных словарях), но также когнитивную, связанную с «картиной мира» говорящего. Когнитивный «паспорт» слова является сверхзадачей словаря и представляет, пожалуй, самую трудоемкую и дискуссионную часть работы над словарем.

Наконец, представленные на страницах словаря энциклопедические справки и прямые ссылки на источники в сети Интернет (словарь является также браузером) позволят читателю ощутить себя частью большого сообщества любителей и знатоков слова, что является предпосылкой для дальнейшего совершенствования словарей и лучшего взаимопроникновения культур в новом тысячелетии.

*Структура страниц*

Отличительная черта страниц словаря – нелинейный способ лемматизации исходного материала. В заголовок вынесена узловая пара слов, например, ЛГАТЬ – ЛЪЖА. Расположенные ниже пары прослеживают цепочку расхождений в значениях членов пары до некоторого предела. Страница условно разделена на две части: под чертой находится справочная база, все над чертой – сопоставительный материал.

*Фиг. 1. Рабочее поле страницы.*

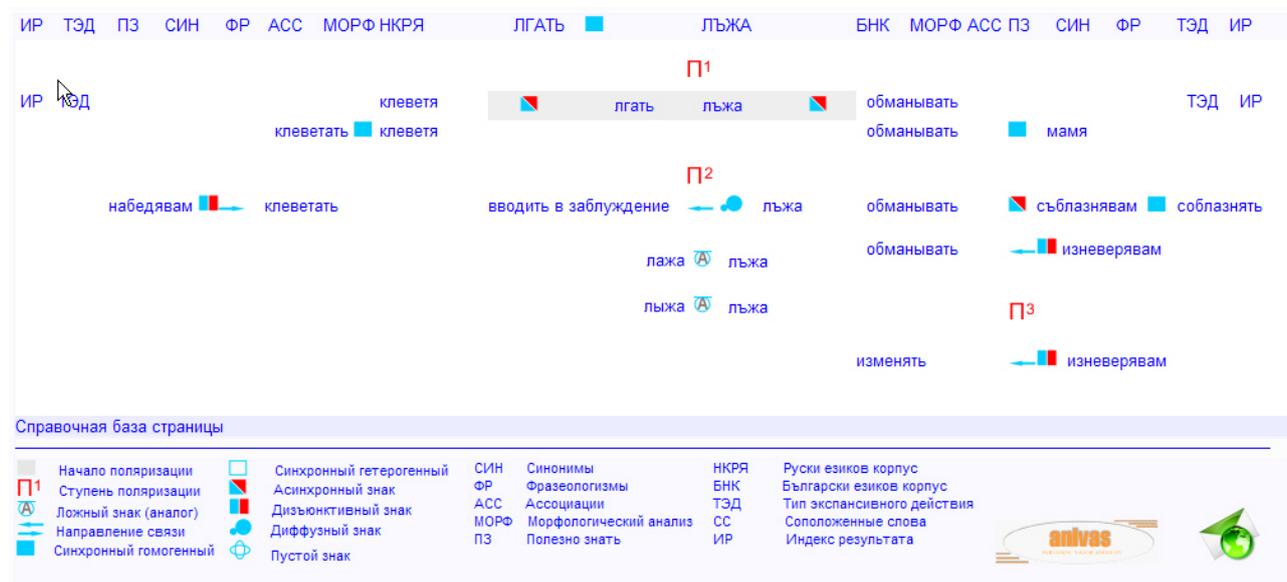

Организация русских и болгарских эквивалентов на странице определяется тем, какое значение слова является первичным по толковому словарю для каждого слова, а какое – вторичным (косвенным). В этом смысле словарь опирается на лексикографические традиции и следует им.

*Фиг. 2. Фрагмент страницы словаря с пояснениями.*

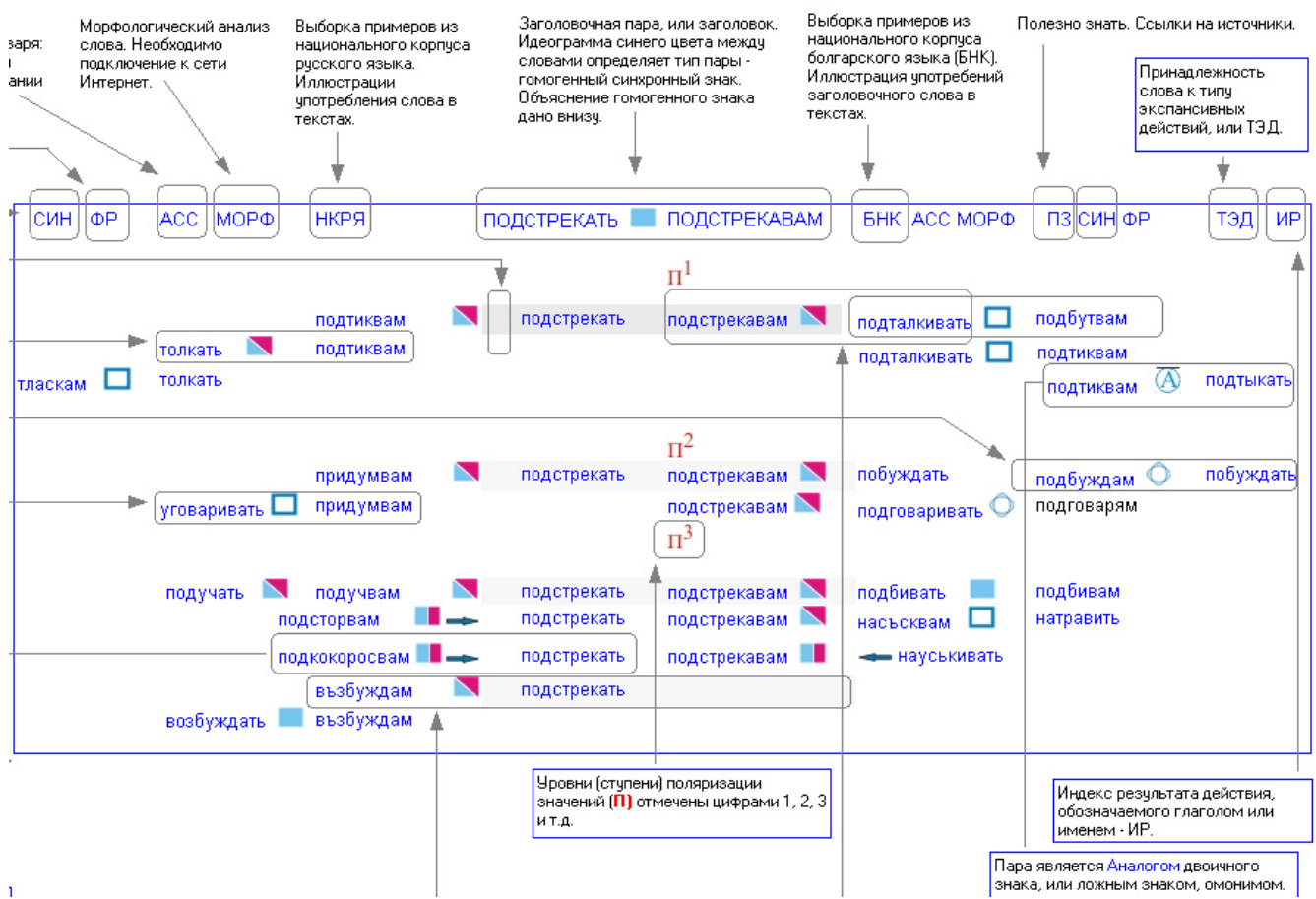

По правилу, словарная страница должна быть в меру избыточной. Учитывая разные стратегии пользователя, целесообразно разбить чтение страницы на строки: каждая строка расширяет семантическое поле словаря и меняет стратегию. В самой же строке расширение семантического поля идет от центра к периферии.

**Строка Первая.** В центре строки находится узловая пара типа лгать – лъжа. Идеограмма ■ (гомогенный синхронный знак) подтверждает общность происхождения и стилистического фона обоих слов. Слева от *лгать* и, соответственно, справа от *лъжа* в плане расширения стратегии пользователя даны ссылки на языковой корпус (НКРЯ и БНК) с примерами и комментариями. Обращается особое внимание на примеры сочетаемости слов в разных контекстах.

*Фиг. 3. Выборка из национального корпуса русского языка.*

Дается также ссылка на: ассоциативный словарь – фрагмент электронной версии словаря для слова в заголовке (АСС), морфологический анализ слов (МОРФ), фразеологический словарь с идиомами, включающими заголовочное слово или передающими аналогичное значение (ФР), синонимы заголовочного слова (СИН).

На этом заканчивается традиционно словарная информация в строке Первой. Далее в строке даны следующие рубрики: полезно знать (ПЗ) – ссылки на источники, энциклопедические сведения и другие материалы по теме; индекс результата действия, обозначаемого глаголом или другим именем (ИР); принадлежность слова к фрагменту схемы (ТЭД). Рубрики (ИР) и (ТЭД) отражают классификацию имен экспансивных действий и концептуальные параметры слова в едином пространстве Словаря.

***Строка Вторая***. Между первой и второй строкой появляется идеограмма П¹ – первый уровень поляризации значений. Вкратце, после П¹ пара распадается, значения слов дифференцируются (для пары многозначных слов) и отображение этих различий развернуто в соположенные знаки.

*Фиг. 4. Соположенные пары клеветя – лгать и лъжа – обманывать.*

Смысл введения второй и следующих строк сводится к следующему.

Пошаговое развертывание различий (дизъюнктивного отношения) в смысловой структуре заголовочной пары. Так, соположенный знак лгать ◤ клеветя реализует значение 'клеветать', характерное только для русского *лгать: лгать на кого-то, Ты лжешь на меня* (СВ *налгать*). С другой стороны, болгарское *лъжа* реализует значение 'вводить в заблуждение' и образует двоичный знак с русским *обманывать*.

Обратной стороной этой пропорции является асинхронный характер двух новых пар, что помечено идеограммой ◤. Это объясняется тем, что *клеветя* и *обманывать* входят в синхронные знаки клеветя клеветать, обманывать мамя и как таковые по рангу должны стоять в заголовках двух самостоятельных словарных страниц. В соответствии с концепцией словаря синхронная связь, если она прослеживается, является первостепенной для организации страниц.

Очень часто при переходе к вторичным значениям заголовочного слова обнаруживается скольжение обозначения к другому типу экспансивного действия, то есть изменяется семантическая протяженность слова в едином пространстве тезаурусной схемы действия. Если читателю достаточна только информация о возможных эквивалентах перевода, эти сведения для него избыточны. Однако стоит отметить, что СЭД классифицирует не только эквиваленты и типы знаков, но также языковое картирование фрагментов тезауруса. Если р. *лгать* в исходном значении 'говорить неправду' отображает семантику дезинформирующих действий и в сознании говорящего ассоциируется с ценностью информации как ориентира, то в значении 'клеветать' глагол примыкает к именам принижающих действий и ассоциируется с дискредитацией субъекта. Адекватное отображение этих различий на Второй строке дано в ТЭД (тип экспансивного действия) слева и справа от глагольных пар.

Соответственно, меняется индекс результата (ИР) для действия, обозначаемого глаголом:

лгать лъжа ИР = (не правда)
лъжа обманывать ИР = (обманут / излъган)
лгать клеветя ИР = (опозорен / опорочен).

***Строка Третья.*** Применительно к словарной странице ЛГАТЬ – ЛЪЖА (см. *фиг.1* выше) это строка: клеветать ■ клеветя -------- обманывать ■ мамя.

Обратите внимание на идеограмму ■, которой помечены глагольные пары. Итак, мы вышли из узловой пары Первой строки, чтобы проследить семантическую поляризацию двоичного знака лгать лъжа на строке Второй и в итоге на строке Третьей круг замкнулся: синхронные ■ пары слов клеветя клеветать, обманывать мамя по определению должны стать предметом самостоятельных страниц словаря. В этом и состоит главное послание строки читателю. Однако на этом информация о заголовочной паре не исчерпывается. Мы переходим к следующему уровню поляризации П².

***Строка Четвертая.*** В эту строку вынесены сведения о диффузии понятий, связанной с уникальностью одного из значений заголовочных слов: вводить в заблуждение ☁ лъжа.

Диффузные ☁ знаки представляют реализованную на межъязыковом уровне возможность простого (однословного) и дефинитивного (с помощью описательных выражений) обозначения. Диффузные пары состоят из уникального слова (значения) в одном из языков и описательного эквивалента, приближающегося по значению к толкованию слова в языке оригинала. Диффузная связь является однонаправленной.

Идеограмма ☁ в паре лъжа вводить в заблуждение, предупреждает, что болгарское *лъжа* имеет, кроме прочих, также развернутый эквивалент, по смыслу выравнивающий данный диффузный знак со знаком лъжа ◣ обманывать. В двоичной модели настоящего словаря, как мы отметили, смысловые диспропорции необходимо описывать с помощью соположенных знаков (пар). Диффузные двоичные знаки не являются простыми: при отсутствии прямого горизонтального соответствия между языками А и В определяющей является вертикальная связь с фрагментами подъязыка.

***Строка Пятая.*** Информация в последней строке имеет предупредительный характер. Здесь представлены прежде всего межъязыковые омонимы и паронимы, т.е. аналоги слов, которые представляют собой ложные ⚠ или пустые знаки: лажа ⚠ лъжа, лыжа ⚠ лъжа. В концепции словаря ложные знаки типа *лажа лъжа* или пустые знаки типа свалить ⊕ свалям 'снимать' не могут стоять в заголовке страницы. Нажатие мышью на *лажа* или *лыжа* откроет толкование слова в языке оригинале. В данном случае словарь дает прямую ссылку на источники в сети.

***Импликативные цепи.*** Как мы отметили, расширение семантического (поискового) поля страницы имеет не только построчный характер, но также направление от центра к периферии для каждой строки. Таким образом формируются ступеньчатые импликативные цепи соположенных знаков типа (см. *фиг. 1*): лгать ... клеветя ... клеветать ... набеждавам; лъжа ... обманывать ... мамя; обманывать ... изневерявам ... изменять.

Мы намеренно отказываемся от прямой нумерации значений многозначных слов в словаре, ограничившись простым повтором слова на странице. Нумерация избыточна, варьирует в зависимости от полноты толковых словарей, и в данном случае ее можно заменить соположенным знаком. Однако мы постарались сделать более прозрачными критерии сопоставимости толкований соответствующих значений, для чего соотносимые толкования значений пары всплывают закрашенными в одинаковый цвет. Например, для пары лгать клеветя при нажатии мышью в интерактивном окне всплывут толкования, маркированные синим цветом.

***Интерактивность.*** К странице предъявлены следующие требования:
1. Задержать внимание читателя, насколько это возможно, в пределах страницы, без прямого выхода на другие с последующим возвратом и возможными при этом ошибками в навигации. Всплывающие окна привязаны локально к интересующему читателя слову и появляются рядом, не перекрывая само слово.
2. Исключить любые ненужные прокрутки страницы. Высота страницы и ее видимость должны обеспечить максимальный комфорт пользователю: эргономика электронного словаря, или «экономия движений руки-мыши», на наш взгляд, уже сейчас является критерием качества словаря. Браузер пользователя может содержать настройки или подключаемые модули, раздувающие панель управления за счет окна эффективного просмотра.
3. Предоставить читателю необходимую краткую справочную базу под чертой во избежание непрерывных отвлекающих пересылок к основному модулю. Контекстуальная помощь обеспечивает минимум сведений относительно типологии двоичных знаков, принятых сокращений и правильного понимания идеограмм.

На странице словаря это выглядит так:

*Фиг. 5. Справочная база страницы.*

| | Справочная база страницы | | | | | | |
|---|---|---|---|---|---|---|---|
| ▢ | Начало поляризации | ☐ | Синхронный гетерогенный | СИН | Синонимы | НКРЯ | Руски езиков корпус |
| П¹ | Ступень поляризации | ◩ | Асинхронный знак | ФР | Фразеологизмы | БНК | Български езиков корпус |
| Ⓐ | Ложный знак (аналог) | ◨ | Дизъюнктивный знак | АСС | Ассоциации | ТЭД | Тип экспансивного действия |
| ⎯ | Направление связи | ● | Диффузный знак | МОРФ | Морфологический анализ | СС | Соположенные слова |
| ■ | Синхронный гомогенный | ✥ | Пустой знак | ПЗ | Полезно знать | ИР | Индекс результата |

***Всплывающие окна (pop-ups).*** Всплывающие окна являются «кадром в кадре». В отличие от своих назойливых собратьев в глобальной сети, мини-окна в словаре являются неотъемлемой частью интерфейса страницы, контекстно привязаны к активному элементу и компрессируют страницу в многослойную структуру. Впечатление «суховатости» страницы при первом приближении исчезнет по мере продвижения к ее глубинным элементам. Соблюдается следующее правило: может быть активным только одно окно. Пользователь не должен бояться нагромождения окон – при открытии нового старое закрывается.

Основное предназначение окон: толкования слов на странице, справка об идеограммах, принятых сокращениях, информация о синонимах, индексы результата (ИР), тезаурусная типология слова (ТЭД), ассоциации слова (АСС), выборки из языковых корпусов для слова, фразеологизмы. Мини-окна служат мостом для перехода к другим страницам, если читатель пожелает расширить смысловое поле поиска. Например:

*Фиг. 6. Мини-окно с толкованием значения глагола и существительного.*

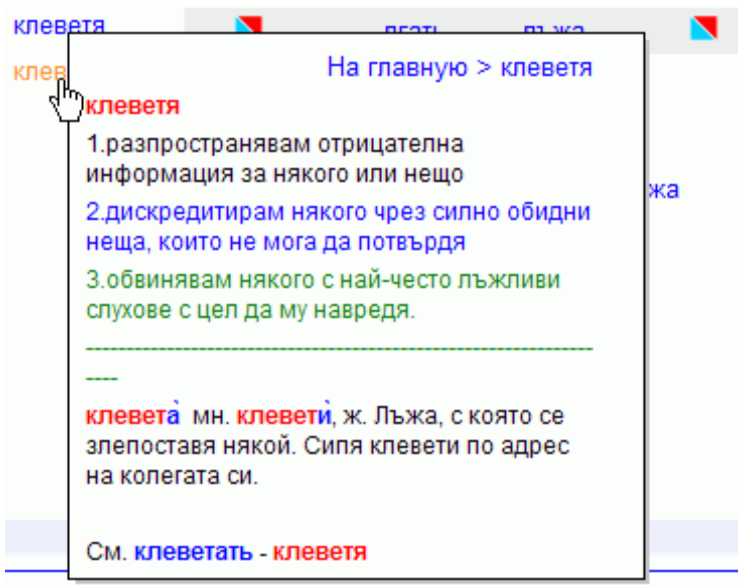

Для статистики приведем пример страницы ЛГАТЬ ЛЪЖА: в первоначальном варианте пользователь мог (не покидая страницы!) открыть 64 мини-окна, а количество сделанных «вручную» файлов «за кадром» страницы превысило 60 документов. Сюда не входят файлы, генерируемые программой.

Раздел «Структура страниц» словаря составлен для ознакомления с принципами организации словаря. Он создан на примере словарной страницы ЛГАТЬ ЛЪЖА и отображает более или менее полно концептуальное ядро словаря. Как и следует ожидать, структура страниц не может быть жестко-однообразной классификационной сеткой в силу различий между типами двоичных знаков.

### *Двоичные знаки*

Как в смысловой структуре отдельно взятого слова наблюдается отмежевание новых значений, так и в структуре двоичного знака фиксируются сдвиги, которые можно рассматривать в двух планах – история и современное (для данного промежутка времени) состояние. Синхронный анализ двоичных знаков фиксирует итоги. Однако граница между историей и итогом является зыбкой (см. позицию Анны А. Зализняк по поводу различий между отношениями семантической производности, связывающими между собой разные значения одного слова на уровне синхронной полисемии, и отношениями между значениями слова в разные моменты его истории [Зализняк 2001]).

Категориальный аппарат словаря состоит из классификации двоичных знаков, достаточно четкой и ясной, чтобы построить систему сопоставления. Система значительно упрощает следующие шаги. В настоящем разделе предложена классификация двоичных знаков, показаны переходы от одних к другим в той мере, в какой это обеспечивает максимальное «прочтение» словарной страницы.

### *Классификация двоичных знаков*

Синхронные г о м о г е н н ы е ■ знаки формируются инерционным путем в обоих языках: это слова общего происхождения, не претерпевшие в исходном значении каких-либо изменений, например, *лгать – лъжа*, *убивать – убивам,* либо заимствованные позже из языка А в язык В типа *грозить – грозя,* либо из третьего языка в языки А и В типа *фальшивить –*

*фалшива*. Критерии определения гомогенного знака: (1) Пара слов имеет общее происхождение, рефлексия общности прослеживается, имеет совпадающее написание (звучание) и одно первичное значение. (2) Пара слов может иметь разные лексикографические интерпретации, но одно первичное значение. (3) Пара слов «непереводима»: она идентична для обоих языков и аутентична для говорящих. (4) Пара слов является «автосинонимом»: смысл, понятие, первичное значение членов пары нерасчленимо в сопоставительной модели тезауруса. (5) Пара слов отсылает к одному и тому же фрагменту тезауруса, имеет один и тот же индекс результата (ИР).

Проверим, достаточно ли этих признаков, чтобы отличить лгать ■ лъжа от следующих пар: лгать клеветя, врать лъжа, обманывать лъжа? Пара лгать клеветя не содержит ни одного признака синхронного гомогенного знака. В значении 'распространять о ком-л. заведомо ложные слухи' эта пара отличается от исходной также индексом результата действия:

     лгать лъжа    ИР (не правда / не истина) – дезинформирующее действие.
     лгать клеветя  ИР (опозорен / опорочен) – принижающее действия.

Пара врать лъжа схожа с первой только по признакам (2) и (5). Признак (3) не может приписываться более чем одной паре слов и является производным от признака (1). Пара обманывать лъжа схожа с первой только по признаку (5). Как видно, пары слов объединяются по первичным или вторичным значениям одного из членов пары, при этом создаются оппозитивные отношения, которые можно выразить в иных пропорциональных знаках:

     лгать клеветя – клеветать ■ клеветя
     обманывать лъжа – обманывать ■ мамя

Для пары врать лъжа это невозможно, поскольку члены пары пропорциональны в первичном значении. Следовательно, перед нами синхронный г е т е р о г е н н ы й знак.

Синхронные г е т е р о г е н н ы е знаки □ формируются параллельным путем в обоих языках. Чаще всего – путем дивергенции или заимствования в А или В из третьего языка, например: *врать – лъжа, вешать – беся, казнить – екзекутирам*. Критерии: (1) Пара слов имеет (но не прослеживается рефлексия общности) или не имеет общего происхождения, написание (звучание) не совпадает, но имеет одно первичное значение. (2) Пара слов может иметь разные лексикографические интерпретации, но одно первичное значение. (3) Пара слов эквивалентна: л е в о е и п р а в о е слово о б о з н а ч а ю т одно и то же в двух языках. (4) Пара слов не является автосинонимом: первичное значение расчленимо в сопоставительной модели тезауруса (беся вешать, казнить екзекутирам). (5) Пара слов отсылает к одному и тому же фрагменту тезауруса, имеет один и тот же индекс результата действия. Проверим, достаточно ли этих признаков, чтобы отличить в словаре вешать беся от вешать веся, с одной стороны, окружать окръжавам от окружать обкръжавам, с другой, казнить екзекутирам и экзекутировать екзекутирам – с третьей.

Пары вешать ■ веся и вешать □ беся соответствуют всем критериям гомогенных и гетерогенных знаков. Пары определяются по значениям (не будем забывать о возможных различиях в лексикографической интерпретации значений):

     (а) вешать ■ веся – рус. 1. Помещать в висячем положении.
                           болг. 1. Окачвам да виси.
     (б) вешать □ беся – рус. 2. Подвергать кого-л. смертной казни через повешение.
                           болг. 1. Умъртвявам чрез окачване на въже, което пристяга
                                   шията и задушава жертвата.

Пара (а) гомогенна, но не отсылает к схеме экспансивного действия, поскольку обозначает нейтральное действие (*вешать белье*). В словаре ей не будет уделена специальная страница. Пара (б) гетерогенна, хотя и восходит к одному этимону – не срабатывает рефлексия общности происхождения (дивергенция языков). Пара отсылает к схеме экспансивного действия, обозначая ликвидирующее по сути действие. Однако как определить соотношение двух пар? То, что они являются соположенными парами, не вызывает сомнения. Отношение

между ними можно назвать а с и н х р о н н ы м  и изобразить схематически следующим образом:

| вешать 1.  веся | или | беся    вешать    веся |
|  | | 1.  1.  1.  2.  2.  2. |
| вешать 2. 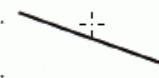 беся 1. | | |

Асинхронное отношение между двумя значениями глагола *вешать* определимо как таковое лишь в сопоставительной модели словаря. Оно имеет причину и следствие. Причина состоит в тенденции развития многозначности (полисемии) слов и в наслоении вторичных значений. Следствие многогранно, но нас будет интересовать в дальнейшем лишь один сопоставительный аспект – нарушение семиологического баланса на языковых полюсах: Подробнее об этом будет сказано ниже. Отметим, что между ситуациями <вешать бельё> и <вешать преступника> существует лишь подобие, но нет онтологической связи. Бельё не может *вешаться* (*бельё повесилось*) ни в первом, ни во втором значениях, но человек может *вешаться на шею кому-нибудь* (к первому значению) и *повеситься*. Если определить вешать 2. как «условный омоним» первому, то кардинально меняется структура взаимосвязей знаков и способ их отображения в словаре. Асинхронной связи уже не будет, баланс восстановлен и выглядит так:

вешать I.  веся
вешать II.  беся

Обратимся к парам окружать окръжавам и окружать обкръжавам. Есть ли смысл рассматривать их в словаре как две самостоятельные пары? Ведь вполне возможен и вариант слово – ряд слов типа *окружать – окръжавам, обкръжавам*. Но в таком случае мы допустим две неточности: отклонимся от принципа двоичного представления слов в сопоставительной модели и обойдём вниманием некоторые ограничения на сочетаемость слов, что имеет прямое отношение к схеме экспансивного действия. Диагностическим тестом для русского *окружать* станет возможность употребления совершенного и несовершенного вида в сочетаниях:

| Значение НСВ (МАС) | Значение СВ (МАС) |
|---|---|
| 2. Находиться, быть расположенным вокруг кого-, чего-л. | 1. Расположиться, занять место вокруг кого-, чего-л. |
| *Деревья окружали дом.* | но не *\*Деревья окружили дом.* |
| *Войска окружали деревню.* | но также *Войска окружили деревню.* |

Далее, подобный тест для болгарского покажет предпочтение глагола *обкръжвам* к активному субъекту действия: 1. Заграждам от всички страни. *Ловците обкръжиха лисицата*. Эти смысловые различия связаны с «картиной мира» носителей языка и подтверждают «пристрастность» пары окружить обкръжавам к схеме блокирующих действий.

Наконец, проанализируем пары казнить екзекутирам и экзекутировать екзекутирам. Здесь – явный лексикографический казус. Пара экзекутировать екзекутирам станет «ложным другом» словаря и переводчика, если вслед за академическими словарями русского языка не признавать факта, что слово *экзекутировать* вошло в обиходную речь. Достаточно пройти по ссылке http://go.mail.ru/, набрав слово в окне поиска. И если в первом значении оно поразительно точно соответствует болгарскому *екзекутирам* 'казнить', то в компьютерном сленге набирает обороты его новый тезка *экзекутировать* от англ. execute – 'выполнить', ср: *Экзекутировать его уже имеют возможность все посетители сервера*. Возможны два варианта:

1. Слово не вошло в широкий обиход, его как бы нет. Тогда казнить образует гетерогенный знак с екзекутирам и ему посвящается страница словаря казнить  екзекутирам.

2. При наличии ряда *экзекуция, экзекутор, экзекуторский* закрепление нового глагола по аналогии с другими рядами в русском предсказуемо и авансом внесено в словарь как страница экзекутировать ■ екзекутирам. Мы склонны отдать предпочтение такому решению.

Синхронные знаки выполняют функцию названия (имени действия, признака или предмета) в обоих языках, не обусловленную и не опосредованную импликативной связью с другими знаками. Отношения между частями синхронного знака являются аналогом логической конъюнкции: Пара лгать клеветя таковой не является. Рус. *лгать* определимо через 'клеветать / клеветя'. Глагол набрал инерцию и сместился частично в зону другого фрагмента тезауруса – принижающие действия. В русском появляются две новые семантические единицы – совершенный вид *налгать* к значению 'клеветать', в отличие от *солгать* к значению 'говорить неправду', и глагол *оболгать* в качестве корректора семантики глагола *клеветать*. Этот сдвиг в обозначении влечет за собой изменение семантического баланса в системе тезаурусная схема – языки А и В. Перестановка семасиологических акцентов в русском меняет также сложившееся статус-кво в двоичной модели тезауруса. Двоичный знак лгать клеветя является выразителем этой диспропорции и имеет асинхронный характер: лгать ◤ клеветя асинхронно парам лгать лъжа и клеветать клеветя, но асинхронно также самому себе в левой и правой части (знак распадается на ведущий и ведомый компоненты). Назовем знаки с таким отношением а с и н х р о н н ы м и ◤ двоичными знаками и рассмотрим их подробнее.

Итак, мы можем добавить к критериям определения синхронных знаков новый признак: необусловленность, неопосредованность назывной функции и безусловный паритет частей знака.

*Асинхронные знаки*

Асинхронные отношения между компонентами двоичного знака являются аналогом логической дизъюнкции: условием дизъюнкции является центробежная тенденция, характерная для каждого из компонентов в отдельности (частное проявление принципа асимметричного дуализма знака, см. [Karcevskij 1929]). Прослеживание ступеней и ветвей дизъюнктивного отношения является одной из основных задач словаря. В концепции СЭД многозначное слово является необходимым и достаточным условием для создания импликативной цепочки двоичных знаков. Специфика словаря – в том, что прослеживаются только импликативные цепочки знаков, имеющих смысловую нагрузку обозначать или отсылать к определенным фрагментам схемы экспансивного действия. Так, импликативная цепь лъжа – обманывать – изневерявам – изменять... не может быть продолжена глаголом *изменям* 'правя друг / делать другим', но его омонимом *изменям* 'преставам да бъда верен / нарушить верность чему-л'.

Асинхронные ◤ знаки в любом случае гетерогенны, но им нельзя дать характеристику гетерогенности в таком виде, в каком мы использовали ее для синхронных гетерогенных знаков. Общность первичных значений как параметр исключается. Чтобы выявить сложность асинхронных отношений, обратимся к оценочной лексике в двух языках.

Проведем обратный лексикографический эксперимент. Дадим толкования некоторых действий и попробуем подобрать к ним слова из двух языков. Искомое понятие должно получить свою словесную форму «как в первый раз». Подборка слов: р. *портить*, б. *развалям*.

(1) Сделать вещь (предмет, пища, одежда и др.) такой, чтобы она была менее или абсолютно непригодна для употребления. *Портить пищу / Развалям сиренето.*

(2) Сделать так, чтобы настроение у кого-то упало слегка или изменилось до противоположного. *Портить настроение / Развалям настроението.*

(3) Сделать так, чтобы ожидания кого-то от выполнения каких-то действий не состоялись в должной мере либо не состоялись вообще. *Портить встречу Нового года / Развалям посрещането на...*

(4) Сделать так, чтобы механизм (аппарат, конструкт) не работал в должной мере или не работал вообще. *Портить игрушку / Развалям играчката.*

(5) Сделать так, чтобы мероприятие не состоялось в должной мере или не состоялось вообще. *Портить праздник / Развалям годежа.*

(6) Сделать так, чтобы тело или составные части тела не работали в должной мере или не работали вообще. *Портить зрение / Развалям си стомаха.*

(7) Сделать так, чтобы поведение кого-то отклонилось от принятой нормы либо изменилось до противоположного. *Портить ребенка / Развалям детето.*

(8) Сделать так, чтобы кто-то в результате по необъяснимой причине (суеверие) заболел. *Портить, наводить на кого-л. порчу / ???*

(9) Сделать так, чтобы что-то распалось на составные части либо превратилось в бесформенную кучу. *??? / Развалям къщата.*

Мы обнаруживаем несоответствие в пунктах (8) и (9). Искомое значение (8) уникально для русского *портить* и так же уникально для болгарского *урочасвам*. Уникальность ставит лексикографа и переводчика перед необходимостью выбора из нескольких возможных вариантов: урочасвам – портить или подвергать сглазу (сглазить) или портить знахарством, заговорами или насылать болезни или портить (с)глазом и т.д. Двоичные знаки, которые образуются с уникальным словом, представляют собой открытое множество. Дизъюнкция является однонаправленной и определяется в сторону от уникального члена двоичного знака. Таким образом, мы вводим в словарь два новых понятия: д и з ъ ю н к т и в н ы й  з н а к и н а п р а в л е н и е ( или ) дизъюнктивной связи. Мы отметили, что диспропорции в структуре двоичного знака можно выразить в иных «пропорциональных знаках». Если такое невозможно, целесообразно говорить о семиологическом дисбалансе в макросистемах A и B, который устраняется на подъязыковом уровне либо на уровне дефиниций, или описательных речевых выражений.

Роль дефиниций в межъязыковом аспекте нельзя недооценивать. Их способность быть посредником в силу меньшей, чем слово, идиоматичности per se наводит на мысль о возможности ставить их в лакуну, образующуюся в случае крайней дизъюнкции. Характерный пример – болгарский глагол *грозя (загрознявам)*. Общая дескрипция 'делать некрасивым' в русском языке не находит опоры в одном определенном предикате. В болгарском ее носителем является глагол *грозя (загрознявам)*. Это обстоятельство не позволяет ему образовывать синхронный знак со строго определенным русским глаголом. Для него можно задать лишь группу асинхронных  знаков:

| | | |
|---|---|---|
| грозя – уродовать (за вычетом эссенциального компонента)<br>грозя – портить (за вычетом компонента 'делать негодным')<br>грозя – обезображивать (за вычетом внутренней формы) | или | 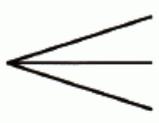 |

Однако ни один из эквивалентов не передает самостоятельно, без нюансов и коннотаций значение 'делать некрасивым'. В данном случае семиологический баланс устанавливается за счет редукции значений компонентов до общей дефиниции типа 'делать некрасивым / правя некрасив'. Обратной стороной этой пропорции является нарочитое исключение некоторых дифференциальных семантических признаков элементов (в).

Вернемся к эксперименту, проведенному выше. Осталось разобраться с пунктом (9): 'сделать так, чтобы что-то распалось на составные части либо превратилось в бесформенную кучу'. Русское *портить* неуместно. Оно находится целиком в зоне оценочных предикатов и не может выражать концептуальную семантику ликвидирующего действия. Между тем, для болгарского *развалям* это значение первично, как первично оно и для русского *разваливать*: рассы́пать, раскидать что-л. собранное в кучу, сложенное. *Развалить бревна. Развалить кучу кирпичей.* □ *Я лежал под разваленной поленницей на куче прошлогодних листьев.* Каверин, Два капитана. || Разг. Сломать, разрушить (постройку, сооружение). *Старик дед сидел у стены*

*разваленной сакли*. Л. Толстой, Хаджи-Мурат. *Снаряд вырыл неглубокую воронку возле колодца, развалив сруб и переломив пополам колодезный журавль*. Шолохов, Тихий Дон (МАС).

Онтология несоответствия между *портить* и *развалям* лежит, таким образом, на пересечении трех слов. Очевидно, асинхронное отношение устанавливается ad hoc (для случая). Если значения обозначить цифрами, то гетерогенные знаки этого типа представляют конъюнкцию исходного (оценочного) значения элемента (a) с вторичным (оценочным) значением элемента (в) при условии, что элемент (в) в своем исходном значении входит в другой гомогенный или гетерогенный знак, например:

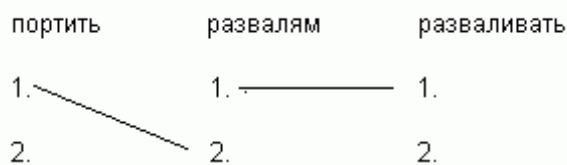

При этом наблюдается скольжение вторичных значений слов в сторону другого фрагмента схемы: для болгарского *развалям* – от ликвидирующего к принижающим действиям (*развалям си мнението за някого*) и к девальвирующим действиям (*развалям хляба* 'понижать качество хлеба'). Знак портить  развалям представляет собой частный случай асинхронных гетерогенных знаков.

### *Диффузные*  *знаки*

Диффузные знаки представляют реализованную на межъязыковом уровне возможность простого (однословного) и дефинитивного (с помощью описательных выражений) обозначения. Диффузные пары состоят из уникального слова в одном из языков и описательного эквивалента, приближающегося по форме к толкованию слова в языке оригинала. Диффузная связь является однонаправленной. Русское *удить* 'ловить удочкой (рыбу)' образует диффузный знак с болгарскими:

удить  *ловя с въдица*
удить  *хващам на въдицата*.

Диффузная связь неопосредованна, предсказуема и синхронна. Семантические пропорции, устанавливаются на уровне дескрипций или общих дефиниций действия. Например, дефиниция 'останавливать, удерживать кого-что-л. каким-н. образом' является общей для следующих глаголов: *хватать, ловить, хващам, ловя; удить, арканить, тралить, треножить, впримчвам*. Мы отмечаем, что для болгарского языка не характерно синкретическое обозначение блокирующего действия и инструмента, в лучшем случае для этого требуется совмещение в имени пространственного и инструментального обозначения (*в-примчвам*). Это продиктовано отсутствием образца. Пропорция возможна лишь в развернутом обозначении, включающем один из глаголов общей группы (удить – *ловя с въдица*, *хващам на въдицата*). Нетрудно увидеть, что в общую дефиницию действия входят слова с отвлеченной семантикой (либо оценочной), что позволяет использовать их в качестве посредников на межъязыковом уровне.

Диффузные знаки не являются простыми. Они состоят из слова и толкующего выражения. Это могло бы создать трудности для составителей сопоставительного словаря, как создавало трудности для составителей переводных словарей, ср. толкование: НЕФТЯ́НИК, -а, м. Работник нефтяной промышленности, а также специалист по нефти (МАС). Перевод: НЕФТЯ́НИК, -а, м. работник в нефтена индустрия; специалист по нефта, петрола (РБР).

Диффузная связь может возникнуть на базе уникальности или оттенка значения слова. Например, лгать и лъжа образуют двоичный знак дезинформирующего действия <говорить что-то, о чем Y знает, что оно не истинно>. Действие замкнуто в сфере X-а, оно безрезультатно, т.к. квалифицируется Y-ом как 'неправда': Y в данном случае – встроенный наблюдатель, выносящий оценку (см. подробнее [Падучева 2007]). Глаголы реализуют одну и ту же семантическую структуру «субъект – действие». Диспропорции в семантической структуре гомогенного знака наступают в тот момент, когда б. *лъжа* вторгается в сферу объекта (сфера Y-а), с этим неизбежно связано обозначение результата (ИР = Y излъган/обманут): *Той ни лъжеше непрекъснато*. Эта диспропорция может быть устранена только включением болгарского переходного *лъжа (някого)* в новый двоичный знак обманывать  лъжа. Но он является асинхронным гетерогенным, поскольку учитывает только горизонтальные связи между языками и оторван от вертикальной связи с концептами схемы, присущей обоим членам гомогенного лгать лъжа. И тут возникает некоторое противоречие. Если б. *лъжа някого* имеет значение 'намеренно ввести кого-л. в заблуждение / мамя, заблуждавам', это значит, что *лъжа* для русскоговорящего имеет значение 'намеренно ввести кого-л. в заблуждение', которое можно выразить словом *обманывать*, но не только этим словом. Если отвлечься от вариаций в толкованиях и придерживаться только правил эквивалентности, то в данном случае эквивалентность *лъжа – обманывать* – вещь отнюдь не очевидная. Это лишь кратчайший путь параметризации смысла по правилу «слово – слово». Удобный, практичный способ избавиться от диффузии понятий в сознании носителя языка. В сопоставительном плане это не всегда корректно. Подчеркнутое «можно выразить» означает вероятность, допущение на фоне дефиниции 'вводить в заблуждение / вкарвам в заблуда'. В качестве компенсации мы предлагаем на странице ЛГАТЬ ЛЪЖА новый диффузный знак лъжа  вводить в заблуждение.

Болгарское *заблуждавам* в словаре Бернштейна (РБС) переводится как *обманывать, вводить в заблуждение*. В нашей классификации знаков это выглядит так:

заблуждавам  вводить в заблуждение (диффузный знак первого уровня, заголовочная пара страницы)
заблуждавам  обманывать ( дизъюнктивный знак второго уровня)
заблуждавам  запутывать ( дизъюнктивный знак третьего уровня, значение 3. по МАС).

*Ложные знаки*  *(аналоги)*

Ложные знаки – общее название для двоичных знаков-омонимов, формирующихся на базе ложных лексических параллелей. Основное отличие ложных знаков от рассмотренных выше типов знаков – потенциальный характер, автоматизм и разрушимость (снятие омонимии толкованием или контекстом). В словаре разрушение омонимии производится путем сопоставления толкований слов. Ложные знаки типа (сущ.) лажа  (глаг.) лъжа, лыжа  лъжа потенциальны в прямой пропорции степени владения неродным языком. Ложные знаки типа грозить  грозя потенциальны в прямой пропорции знанию или «обкатке» заимствований в родном языке из языка А, и комплементарны гомогенным знакам типа грозить  (заимств.) грозя. Ложные знаки типа бранить  браня 'защищать' потенциальны в силу этимологической рефлексии. Ложные знаки типа забаллотировать  балотирам потенциальны в силу аналогизации функций приставки:

баллотировать  балотирам (гомогенный)
забаллотировать  балотирам (ложный)
забаллотировать забалотирам (ложный)
забаллотировать проваля в избори (дизъюнктивный знак)

Ложные знаки в словаре не оцениваются с позиции вредности – полезности данного явления либо случайности – закономерности возникновения.

## *Пустые знаки*

Цитируем не ради смеха: *Ай ем българче. Обичам наште маунтинс зелени. Българче да се наричам – ит из фърстли джой фор мене.* Пустая комбинация болгарского и английского срабатывает и вызывает улыбку. Немаловажное примечание: пустая комбинация значима. *Наште маунтинс* эквивалентна *ауър планини*. Но на каком уровне? Скорее всего – на ассоциативном уровне билингва. Ассоциации в языке могут порождать новые значения, однако становятся также источником помех в общении. Если ассоциации имеют место быть, в равной степени возможны двоичные ассоциативные знаки, создающие «радарные помехи» в восприятии значения. Такие ассоциативные конструкты мы называем п у с т ы м и д в о и ч н ы м и знаками. В сопоставительном исследовании мы определили бы их как межъязыковые паронимы.

Пустые знаки в отличие от ложных имеют непотенциальный характер, семантически заданы кодом и неразрушимы: ср. рус. *напоследок* 'под конец' и болг. *напоследък* 'в последно време / в последнее время'; рус. *заключить* 'лишить свободы' и болг. *заключа* 'оставя в помещение, затворено с ключ / оставить в помещении, закрытом на ключ'; рус. *посягать* 'попытаться завладеть кем-, чем-л.' и болг. *посягам* 'протягам ръка за някакво действие / протягивать руку'. Отнюдь не считая понятие пустого знака неуязвимым, мы хотели бы провести разграничение в чисто практических целях, допустив, что пустые знаки все-таки «полнее» ложных. Непотенциальный характер пустого знака означает отсутствие какого-либо случайного элемента, например, нарочитого уподобления, заимствования или фонетической конвергенции слов (типа рус. *лук* I. и *лук* II.). Семантическая заданность пустого знака определяется общими моделями словообразования: это случаи, когда значение продуктивной модели может довлеть (иметь сильный лексико-семантический потенциал) над значением слова либо находиться с ним в неравноправном симбиозе (заключить заключа). Неразрушимый характер пустого знака определяется тем, что его можно разложить на разные семантические части не простой подстановкой контекста, а специальным семантическим анализом, который, при всей глубине и точности, покажет живую связь между членами знака, ср.:

*Расцепить* | -цеплю, -целишь; -цепле-нный; сов., кого-что. Разъединить (сцепленное, сцепившихся). Р. вагоны. Р. драчунов. || несов. расцеплять, -яю, -яешь. 1Г сущ. расцепка,-и, ас. (спец.) [СО];

*Разцепвам* разцепваш, несв. и разцепя, св.; какво.
1. Разделям на две (чрез цепене). Разцепих дървото. Разцепих навалицата.
2. Пуквам, спуквам, пропуквам, пръскам. Разцепих динята, защото я изпуснах.
3. Прен. За партия, организация и др. под. – разделям, разединявам, разлъчвам.

В паре заключить заключа словообразовательные модели расходятся с типизацией действий. Одинаковая для двух языков модель *за-[ключ]-ать – за-[ключ]-вам* оказалась в двойственном отношении к схеме блокирующих действий. В русском объектом действия является только Y (*заключить под стражу*), в остальных случаях говорят *закрыть на ключ* (дверь), *запереть на замок* (комнату). Болг. *заключвам* в прямом значении связывается с помещением, в которое не должен проникнуть посторонний (одна цель действия), а в косвенном – с Y-ом, который не должен покидать помещения (другая цель действия). В русском элемент *ключ* подвергся десемантизации (ср. *заключить в объятия – обнять, заключить в скобки*), но благодаря транспозиции исходного значения образовано существительное *заключенный*. В болгарском такая транспозиция невозможна (*затворник*) в силу указанной выше специфики. В итоге на межъязыковом уровне глаголы *заключать* и

*заключвам*, хотя и отсылают к одному типу действий, не могут образовывать гомогенного знака, что отмечено в переводных словарях. Для перевода используются глаголы-посредники с блокирующим значением: *заключвам* – запирать, закрывать на ключ; *заключать* – затварям, задържам (в затвор).

Ложные двоичные знаки, а также пустые двоичные знаки, должны служить предостережением для неопытного читателя или начинающего переводчика. В таких случаях рекомендуется внимательное прочтение словарных толкований.

### *Иерархия, переходы знаков и импликативные цепи*

Классификация двоичных знаков является первостепенной задачей на этапе предварительной подготовки словаря. На страницах словаря классификация в сжатом виде представлена идеограммой знака. Главное на страницах – визуализация переходов от одних знаков к другим. Прослеживаются некоторые закономерности, которые интересны в научном плане и полезны в прикладном аспекте.

Синхронные гомогенные ▮ знаки могут стоять в заголовке словарной страницы. Поиск *лгать* или *лъжа* выведет на страницу ЛГАТЬ ЛЪЖА. Синхронные гетерогенные ▯ знаки могут стоять в заголовке словарной страницы. Поиск *вешать* или *беся* выведет на страницу ВЕШАТЬ БЕСЯ.

Асинхронные знаки типа лгать клеветя по рангу не могут образовывать страницы. Их можно обнаружить либо на странице ЛГАТЬ ЛЪЖА, либо на странице КЛЕВЕТАТЬ КЛЕВЕТЯ, на второстепенной строке соположенных знаков. Закономерность: эти асинхронные знаки являются выразителями семантической поляризации заголовочной пары и по правилу заканчивают импликативную цепь новой синхронной парой.

Дизъюнктивные ▮▮ знаки могут стоять в заголовке страницы. Поскольку это знаки с односторонней связью уникального слова и неуникальных слов (загрознявам ⟶ b1, b2, b3), уникальное слово самостоятельно образует страницу. Поиск *портить* приведет к странице ПОРТИТЬ ИЗПОРТВАМ, но поиск *загрознявам* – к странице ЗАГРОЗНЯВАМ ▮▮. Закономерность: открытые множества знаков на первом уровне не имеют внутренней иерархии.

Диффузные ☁ знаки могут стоять в заголовке словарной страницы. Поиск *заблуждавам* приведет нас к странице ЗАБЛУЖДАВАМ ☁ Вводить в заблуждение. По степени важности на первом месте – уникальное слово и толкующий эквивалент. Закономерность: все пары, кроме заголовочной, являются дизъюнктивными.

Ложные знаки (аналоги) и пустые знаки не могут образовывать страницы. По рангу они занимают самые нижние места в иерархии.

Как видно, объединение слов в двоичные знаки может быть «произвольным» и соответствовать разным установкам. Установка на паритетность и гомогенность объединит пару портить ▮ изпортвам. Установка на переводимость объединит портить ◣ развалям в асинхронный знак, но при этом элиминируются различия в первичных значениях. Установка на «закономерные соответствия» уникальных элементов создаст дизъюнктивные оппозиции типа портить ▮▮ урочасвам, урочасвам ▮▮ сглазить, но по закономерности сюда войдут также неуникальные гетерогенные пары типа вешать ▯ беся, казнить ▯ екзекутирам. Установка на толкование создаст знаки-перифразы типа заблуждавам ☁ вводить в заблуждение, злоумышлять ☁ мисля зло някому. Установка на адекватные замены приведет к семантическому перифразированию или интерпретациям, например: кокнуть ▮▮⟶чукна, кокнуть ▮▮⟶счупя, кокнуть ▮▮⟶убия.

Построение системы двоичных знаков фундаментально для словаря. Определение приоритетности знаков первостепенно для организации страниц и системы поиска в словаре. Приоритеты знаков важны для установления узла импликативной цепи, но также для фиксации конца этой цепи на странице.

Итак, на страницах словаря возможны следующие импликативные цепи:

1. 🟦 ------ 🟥 ------ 🟦  Пример:

        *лгать* 🟦 *лъжа*
      *клеветя* 🟥 *лгать*
    *клеветать* 🟦 *клеветя*
  ---------------------- Выход из цепи на новую страницу КЛЕВЕТАТЬ КЛЕВЕТЯ.

2. ⬜ ------ 🟥 ------ 🟦  Пример:

       *стукнуть* ⬜ *чукна*
        *чукна* 🟥 *ударить*
         *ударить* 🟦 *ударя*
  ------------------ Выход из цепи на новую страницу УДАРИТЬ УДАРЯ.

3. 🟦🟥 ----- ⬜  Пример: *кокнуть* ----- *счупя* ----- *сломать*.
  ------------------ Выход на страницу СЛОМАТЬ СЧУПЯ.

4. 🟦🟥 ----- 🟦  Пример: *кокнуть* ------ *убия* ------ *убить*
  ------------------ Выход на страницу УБИТЬ УБИЯ.

5. 🔵 ------ 🟥🟦 ------ 🟦  Пример:

       *заблуждавам* 🔵 *вводить в заблуждение*
     *обманывать* 🟥🟦 *заблуждавам*
   *мамя* 🟦 *обманывать*
  ------------------ Выход из цепи на новую страницу ОБМАНЫВАТЬ МАМЯ.

6. 🟦 ------ 🅰  Пример: *лгать* 🟦 *лъжа*
          *лъжа* 🅰 *лажа* (ложный знак)

7. 🟦 ------ 🔵  Пример:
       *лгать* 🟦 *лъжа*
       *лъжа* 🔵 *вводить в заблуждение*

8. 🟦 ------ ✦  Пример:
      *изменям* 🟦 *изменять*
       *изменять* ✦ *изменям* (пустой знак)

9. 🟦 …… 🟥 …… ⬜  Пример:
      *портить* 🟦 *изпортвам*
        🟥
  *ухудшать* ⬜ *влошавам*
  -------------- Выход из цепи на новую страницу УХУДШАТЬ ВЛОШАВАМ.

  Очевидно, цепи могут наслаиваться, и на странице словаря показ разветвления цепей должен иметь разумный предел. В таких случаях вводится гипертекстовый мост.

### *Идеограммы*

  Идеограмма в общем случае представляет собой значок, картинку и может являться элементом графического интерфейса пользователя в данном электронном приложении. Смысловая нагрузка идеограммы варьирует от простого индикатора до классификатора материала. Идеограммам отведена существенная роль. Они применяются в каждой словарной странице, поэтому в основном модуле словаря им посвящена рубрика «Идеограммы», которая знакомит читателя с основными принципами их размещения и прочтения. Как внешний вид

(форма, фигура, цвет), так и место вставки идеограммы, подчиняются логике сопоставительного словаря.

*Навигация и поиск*

Фреймовая структура словаря делит страницу на две части: основной модуль и рабочее поле. В основном модуле рубрики словаря организованы в иерархическую структуру. Поиск терминов осуществляется при входе в глоссарий. Быстрый комбинированный поиск слов – через поисковый индекс «Поиск». Алфавитный поиск – в рубрике «Навигация и поиск». Комбинированный поиск регулируем: можно задать число соответствий на одну страницу поиска. Например, комбинированный поиск слова *лгать* выведет читателя одновременно на разные тематические рубежи – основная страница, встречаемость в текстах словаря, выборки из национального корпуса, соположенные пары слов, типология экспансивных действий, каталог линков и т.д. Алфавитный поиск дается отдельно для двух языков. Полное описание правил навигации возможно только при достижении словарем некоторого критического объема страниц. По ходу действия не обойтись также без коррекции других компонентов словаря. Описанные выше приоритеты знаков частично помогут избежать недоразумения. Порядок таков: общий алфавитный список болгарских слов и такой же список русских слов. В каждом списке устанавливается алфавитная иерархия: глаголы, существительные, прилагательные, наречия. Если слова нет в алфавитном списке, оно либо не входит в классификацию имен экспансивных действий, либо является фантомом: межъязыковые омонимы и паронимы, отсутствуют в алфавитном списке, но есть на страницах словаря и образуют пустые и ложные знаки.

Поиск по слову выводит на страницу, в которой слово либо образует заголовочную пару с болгарским (гомогенную, гетерогенную или диффузную), либо вынесено самостоятельно в заголовок в силу своей уникальности (дизъюнктивные знаки). Например:

| | | |
|---|---|---|
| АРЕСТОВАТЬ | → | АРЕСТОВАТЬ 🟦 АРЕСТУВАМ |
| ВЕШАТЬ | → | ВЕШАТЬ ⬜ БЕСЯ |
| АРТОБСТРЕЛ | → | АРТОБСТРЕЛ 🔵 Артилерийски обстрел |
| АРТАЧИТЬСЯ | → | АРТАЧИТЬСЯ 🟥🟦 |

Если не удается выйти на страницу, она либо не готова, либо слово не имеет нужного приоритета. Например, омоним *браня* к русскому *бранить* не фигурирует в алфавитном списке. Необходимо воспользоваться комбинированным поиском: в этом случае поиск *браня* выведет на страницу БРАНИТЬ 🟥🟦. Можно воспользоваться комбинированным поиском для получения полной информации о слове в словаре.

Правило синхронности первичных значений слов является определяющим как для организации страниц, так и для поиска. Читатель, может быть, щелкнув по слову *портить* в алфавитном списке, будет надеяться выйти на страницу типа ПОРТИТЬ РАЗВАЛЯМ, по крайней мере, будет считать болгарское *развалям* наиболее вероятным претендентом на узловую пару заголовка. Это справедливо, если учитывать только статистику переводов, но создало бы огромные трудности для организации интегрального словаря. В концепции словаря русское *портить*, будучи оценочным словом и имея оценочного напарника в лице *портя* (*изпортвам*), образует гомогенную пару <span style="color:blue">портить изпортвам</span> и выносится в заголовок.

Правило синхронности заимствований не всегда согласуется с правилом синхронности первичных значений. Поиск болг. *ангажирам* откроет страницу АНГАЖИРОВАТЬ АНГАЖИРАМ. Это не значит, что не учитываются особенности употребления и удельный вес каждого слова в родном языке. Глагол *ангажирам* в переводных словарях имеет эквиваленты *предлагать, приглашать* [БРС], *нанимать, заказывать, бронировать*. При таком многообразии эквивалентов перевода именование страницы ставит перед необходимостью выбора и предпочтения. Предполагая, что содержание страницы обеспечивает необходимые сведения о коррелятах, мы считаем название АНГАЖИРОВАТЬ АНГАЖИРАМ технически обоснованным.

Правило синхронности метонимического образования реализовано в узле страницы ГЛАВАРЬ ГЛАВАТАР(КА). Пометы *уст., разг., неодобрит.* несущественны для объединения пары в двоичный знак. Нюансы значения слова в сочетаниях (*главарь мятежа* 'зачинщик', *главарь банды* 'вожак', *главарь мафии* 'руководитель', *главарь индейцев* 'вождь') несущественны для заголовка. Перевод болгарского *главатар* – 1) лидер, предводитель; 2) уст. атаман, главарь [БРС] – включает неудачное заимств. *лидер* и реалию *атаман*. Можно перевести *главарь* как *тартор*, но их нельзя поставить рядом в заголовке страницы.

Правило семантико-стилистической адекватности применяется в случаях типа рус. *стыдить – срамить* и болг. *засрамвам – срамя*. Стыдить – к человеку (чувство стыда): *Постыдился бы! Засрами се!* Срамить – к позору, бесчестью, т.е. к социальному статусу. Страницы: СТЫДИТЬ ЗАСРАМВАМ и СРАМИТЬ СРАМЯ.

Правило селекции значений применяется при составлении страницы, в которой только одно из слов узловой пары вошло в орбиту имен экспансивных действий. Таков случай р. *ахнуть* и болг. *ахна*. В общей классификации двоичных знаков они нашли бы отражение как гомогенный знак ахнуть ■ ахна. Но русское *ахнуть* означает также: *перех. Прост.* С силой ударить. [*Мирон Григорьевич*], *почти не размахиваясь, ахнул его по скуле*. Шолохов, Тихий Дон [МАС]. Условимся, что заголовочные слова страницы, которые образуют двоичный знак, но нейтральны схеме экспансивного действия, даны в круглых скобках: АХНУТЬ ■ (АХНА).

Если встречаются дублетные пары типа блюдолиз / лизоблюд – блюдолизец, заголовок страницы оформлен так: БЛЮДОЛИЗ ■ БЛЮДОЛИЗЕЦ □ ЛИЗОБЛЮД. Если встречаются дублеты (уст.) бременить, обременять – обременявам, устаревшее слово дается в заголовке в квадратных скобках: [БРЕМЕНИТЬ] ОБРЕМЕНЯТЬ ■ ОБРЕМЕНЯВАМ.

*Заключение*

Попытка объединить в одном словаре специальный лингвистический интерес, когнитивный взгляд на лексику двух славянских языков и прикладной аспект в некотором смысле необычна и дискуссионна. Реализация подобного словаря может стать плодом коллективных усилий и должна вписываться в бюджет создателей или поручителей проекта.

В проекте словаря практически реализовано трехугольное tertium comarationis: тезаурусная схема – русский язык – болгарский язык. Как справедливо отметил Д. Митев, «важно, чтобы исходная база позволяла установление сопоставимости языков как на поверхностном, так и на глубинном уровне языковых единиц, идентификацию соотносительных единиц разных языков и определение форм проявления межъязыкового соответствия (в пределах от полного тождества до полного различия)» [Митев 2006: 7]. Сфера приложимости используемого в словаре t.c. – лексические системы двух родственных славянских языков. Предложенная классификация эквивалентов является всего лишь одной из возможных для родственных языков. Представляется, что любой интегральный словарь подобного типа не мог бы обладать стройной логической структурой без ясного совмещенного портрета сопоставляемых языков.

Действующий прототип словаря, включающий двадцать типовых страниц (статей), опубликован на сайте составителя: http://anivas32.narod.ru/.

*Keywords*

Схема экспансивного действия, двоичная модель тезауруса, двоичные знаки, синхронные знаки, асинхронные знаки, дизъюнктивные знаки, диффузные знаки, идеограммы знаков, имплицитные аспекты номинации, семантическая поляризация фрагментов тезаурусной схемы в языках.

*Аннотация*

Статья представляет концепцию словаря «Сопоставительный словарь имен экспансивных действий в русском и болгарском языках». Приводятся основные параметры нового интернет-ориентированного сопоставительного словаря, показаны принципы его формирования, классифицированы первостепенные связи между словами-эквивалентами. Показаны принципиальные различия между переводными словарями и моделью интегрального сопоставления. Предложена схема для классификации страниц. Введены новые понятия и ключевые слова. Опубликован реальный прототип словаря и двадцать ключевых страниц. Предлагается для широкого обсуждения возможность исползовать этот прототип в качестве варианта сопоставительного словаря близкородственных языков нового поколения.

*Abstract*

The article is a fragment from a new comparing dictionary "The Comparing Dictionary of the Names of Expansive Actions in Russian and Bulgarian Languages." The main parameters of the new web-orientated comparing dictionary are exposed, the principles of its formation are shown and the paramount connections between the words-matches are classified. The principle difference between the translational dictionaries and the model of double comparison is also shown. The scheme for the page classification is suggested. New notions and key words are inserted. The real prototype of a dictionary with several key pages is published. The probability of this prototype to turn into the version of the Russian-Bulgarian Comparing Dictionary of a new generation is offered for the wide discussion.

**ЛИТЕРАТУРА**

http://www.gramota.ru Официальный сайт: ГРАМОТА.РУ.
http://lexicograf.ru/ Официальный сайт: «Lexicograph».
http://www.ruscorpora.ru/ Официальный сайт: Национальный корпус русского языка.
http://feb-web.ru/ Официальный сайт: ФЭБ «Русская литература и фольклор».
http://www.belb.net/index.php Официальный сайт: БЕЛБ (14/01/10).
http://framenet.icsi.berkeley.edu/ Официальный сайт: FrameNet (08/01/10).
http://www.ibl.bas.bg/BGNC_classific_bg.htm Официальный сайт: Български национален корпус.

Зализняк 2001 – *Зализняк Анна А.* Семантическая деривация в синхронии и диахронии: проект «Каталога семантических переходов» // Вопросы языкознания, 2001, № 2.

Кошелев 2004 – *Кошелев А.Д.* Статьи по лингвистике и когнитивной психологии [Электронный ресурс] // Языки славянских культур [сайт]. 1993 – http://www.lrc-press.ru/05.htm (20/08/09).

Легурска 2002 – *Легурска П.* Анализ на предметните имена в руския и българския език. [Электронный ресурс] // БЕЛБ [сайт]. 2009 – http://www.belb.net/projects/proektpm/VTOR.htm (14/01/10).

Липовска 2006 – *Липовска А.* Некоторые аспекты работы над семантическим модулем комплексного русско-болгарского словаря // Аспекты контрастивного описания русского и болгарского языков. Шумен, 2006.

Липовска 2009 – *Липовска А.* Русско-болгарская лексикография: традиции и тенденции развития. София: Изд-во Софийского университета, 2009.

Митев 2006 – *Митев Д.* К вопросу о теоретико-методологических основах сопоставительной болгарско-русской грамматики // Аспекты контрастивного описания русского и болгарского языков. Шумен, 2006.

Падучева 2007 – *Падучева Е.В.* Когнитивные идеи в теоретической семантике [электронный ресурс] // The project «Lexicograph» [сайт]. М.: 2005 – http://lexicograf.ru/files/cognit_2007_fin.pdf (08/10/08).

Паликова 2007 – *Паликова О.* Двуязычный словарь и функционально значимые связи слова: Дисс. … докт. филос. по рус. языку. Тарту, 2007.

Перванов 1995 – *Перванов Я.А.* Имплицитные аспекты номинации в русском и болгарском языках: Дисс. … канд. филол. наук. Одесса, 1995.

Перванов 2009 – *Перванов Я.А.* Семантическая поляризация экспансивного действия в русском и болгарском языках // Болгарская русистика, 2009, № 2.

Перванов 2010 – *Перванов Я.А.* Схема экспансивного действия. Русские глаголы [Электронный ресурс] // Anivas [сайт]. 2009 – http://anivas32.narod.ru/sed_out/sxema_dejstwia.htm

Розина 2003 – *Розина Р.И.* Динамическая модель семантики глагола ВЗЯТЬ // Русский язык сегодня. Вып. 2. М.: Азбуковник, 2003.

Скаличка 1989 – *Скаличка В.* Типология и сопоставительная лингвистика. // Новое в зарубежной лингвистике // Вып. XXV Контрастивная лингвистика. М.: Прогресс, 1989.

Табанакова, Ковязина 2007 – *Табанакова В.Д., Ковязина М.А.* Новая модель двуязычного экологического словаря-тезауруса // Актуальные проблемы лингвистики и терминоведения. Урал. гос. пед. ун-т; Институт иностранных языков. – Екатеринбург, 2007.

Филлмор 1983 – *Филлмор Ч. Дж.* Об организации семантической информации в словаре // Новое в зарубежной лингвистике. Вып. XIV Проблемы и методы лексикографии . М.: Прогресс, 1983.

Fillmore, Baker, Sato 2002 – *Fillmore, Charles J., Baker, Collin F. and Sato, Hiroaki* (2002): Seeing Arguments through Transparent Structures // In Proceedings of the Third International Conference on Language Resources and Evaluation (LREC). Las Palmas. 2002.

Karcevskij 1929 – *Karcevskij S.* Du dualisme asymetrique du signe linguistique // Travaux du cercle linguistique de Prague, 1929, № 1.

Nida 1964 – *Nida E.* Toward a Science of Translating: With Special Reference to Principles and Procedures Involved in Bible Translating, Leiden, E. J. Brill, 1964.

**В работе над словарем использованы:**

СО 1984 – Словарь Ожегова / Под ред. Н.Ю. Шведовой. М.: Русский язык, 1984.

РБЕ 1977-1987 – Речник на българския език. София: Изд-во на БАН, 1977-1987.

БРС – Болгарско-русский словарь. Ок. 58000 слов. / Сост. проф. С.Б. Бернштейн. М.: Советская энциклопедия, 1966.

МАС – Словарь русского языка в 4-х томах М.: Русский язык, 1999.

Даль – Даль В. Толковый словарь живого великорусского языка (современное написание слов). М.: Цитадель, 1998 г.

Зализняк – Зализняк А.А. Грамматический словарь русского языка. 3-е изд. Словоизменение. М.: 1987.

РАСС – Русский ассоциативный словарь: электронная версия. [Электронный ресурс] // http://tesaurus.ru/dict/dict.php (11/08/09).

Абрамов – Абрамов Н. Словарь русских синонимов и сходных по смыслу выражений. М.: Русские словари, 1999.

РБР – Руско-български речник: Русско-болгарский словарь / Под ред. на С.Влахов и Г.А. Тагамлицка: В 2 т. – София: Наука и изкуство, 1985-1986. Т.1 – 2.

Димитрова, Спасова – Димитрова М., Спасова А. Синонимен речник на съвременния български книжовен език. София: Изд-во на БАН, 1980.

РЧД – Речник на чуждите думи в българския език, София: Наука и изкуство, 1999.